# Neural-based machine translation for medical text domain. Based on European Medicines Agency leaflet texts.

Krzysztof Wołk, Krzysztof Marasek
kwolk@pja.edu.pl
*Department of Multimedia, Polish Japanese Academy of Information Technology, Koszykowa 86, 02-008 Warsaw, Poland*

**Abstract**

The quality of machine translation is rapidly evolving. Today one can find several machine translation systems on the web that provide reasonable translations, although the systems are not perfect. In some specific domains, the quality may decrease. A recently proposed approach to this domain is neural machine translation. It aims at building a jointly-tuned single neural network that maximizes translation performance, a very different approach from traditional statistical machine translation. Recently proposed neural machine translation models often belong to the encoder-decoder family in which a source sentence is encoded into a fixed length vector that is, in turn, decoded to generate a translation. The present research examines the effects of different training methods on a Polish-English Machine Translation system used for medical data. The European Medicines Agency parallel text corpus was used as the basis for training of neural and statistical network-based translation systems. The main machine translation evaluation metrics have also been used in analysis of the systems. A comparison and implementation of a real-time medical translator is the main focus of our experiments.

*Keywords:* machine translation; statistical machine translation; nlp; text processing

## 1. Introduction

Machine Translation (MT) is a computer's translation of text with no human assistance. MT systems have no knowledge of language rules. Instead, they translate by analyzing large amounts of text in language pairs. They can be trained for specific domains or applications using additional data germane to a selected domain. MT systems typically deliver translations that sound fluent, although they tend to be less consistent than human translations. Statistical machine translation (SMT) has rapidly evolved in recent years. However, existing SMT systems are far from perfect, and their quality decreases significantly in specific domains. The scientific community has been engaged in SMT research. Among the greatest advantages of statistical machine translation is that perfect translations are not required for most applications [1]. Users are generally interested in obtaining a rough idea of a text's topic or what it means. However, some applications require much more than this. For example, the beauty and correctness of writing may not be important in the medical field, but the adequacy and precision of the translated message is very important. A communication or translation error between a patient and a physician in regard to a diagnosis may have serious consequences on the patient's health. Progress in SMT research has recently slowed down. As a result, new translation methods are needed. Neural networks provide a promising approach for translation [2].

Machine translation has been applied to the medical domain due to the recent growth in interest in and success of language technologies. As an example, a study was done on local and national public health websites in the USA with an analysis of the feasibility of edited machine translations for health promotional documents [3]. It was previously assumed that machine translation was not able to deliver high quality documents that can used for official purposes. However, language technologies have been steadily advancing in quality. In the not-too-distant future, we expect that machine translation will be capable of translating any text in any domain at the required quality.

In our opinion, the medical data field is a bit narrow, but very relevant and a promising research area for language technologies. Medical records can be translated by use of machine translation systems. Access to translations of a foreign patient's medical data might even save their life. Direct speech-to-speech translation systems are also possible. An automated speech recognition (ASR) system can be used to recognize a foreign patient's speech. After it is recognized, the speech could be translated into another language with synthesis in real time. As an example, the EU-BRIDGE project intends to develop automatic transcription and translation technology. The project desires innovative multimedia translation services for audiovisual materials between European and non-European languages [http://www.eu-bridge.eu].

Making medical information understandable is relevant to both physicians and patients [4]. As an example, Healthcare Technologies for the World Traveler emphasizes that a foreign patient may need a description and explanation of their diagnosis, along with a related and comprehensive set of information. In most countries, residents and immigrants communicate in languages other than the official one [5].

Karliner et al.[6] talks of the necessity of human translators obtaining access to healthcare information and, in turn, improving its quality. However, telemedicine information translators are not often available for either professionals or patients. Machine translation should be further developed to reduce the cost associated with medical translation [7]. In addition, it is important to increase its availability and overall quality.

Patients, medical professionals, and researchers need adequate access to the telemedicine information that is abundant on the web [8]. This information has the potential to improve our health and well-being. Medical research could also improve the sharing of medical information. English is the most used language in medical science, though not the only one.

Due to its complexity, Polish is considered to be one of the most challenging West-Slavic languages. This makes Polish [9] translation very difficult for an SMT system. Polish grammar includes the complications of language elements and rules, along with an immense vocabulary. Word order in sentences is also a problem. In addition, the language has seven cases and fifteen gender forms of nouns and adjectives.

The goal of this paper is to present our experiments on neural based machine translation in comparison to statistical machine translation. The adaptation of translation techniques as well as proper data preparation for the need of PL-EN translation was also necessary. Such systems could possibly be used in real time speech to speech translation systems and aid foreign travelers that would require medical assistance. Combining such system with OCR and augmented reality tool could bring to mobile devices a real time translator as well. Human interpreters with a proper medical training are extremely rare and costly. Machine translation could also assist in the evaluation of medical history, diagnoses, proper medical treatments, health related information, and the findings of medical researchers from entire word. Mobile devices, the Internet and web applications can be used to boost delivery of machine translation services for medical purposes even as real time speech-to-speech services.

The article is structured as follows. The Section 2 describes main concepts of neural network usage for machine translation. The Section 3 gives some background regarding the data that was used in the experiments and its preparation. The Section 4 contains brief descriptions of evaluation metrics that were used. In the Section 5 the technical aspects of both SMT and neural based translation systems are described. Finally the Section 6 presets our results and conclusions.

## 2. Neural networks in translation

Machine learning is the programming of computers for optimization of a performance criterion with the use of past experience/example data. It focuses on the learning part of conventional intelligence. Based on the input type available while training or the desired outcome of the algorithms, machine learning algorithms can be organized into different categories, e.g., reinforced learning, supervised learning, semi-supervised learning, unsupervised learning, development learning, and transductive inference. With a number of algorithms, learning of each type can take place. The artificial neural network (ANN) is a unique learning algorithm inspired by the functional aspects and structure of the brain's biological neural networks. With use of ANN, it is possible to execute a number of tasks, such as classification, clustering, and prediction, using machine learning techniques like supervised or reinforced learning. Therefore, ANN is a subset of machine learning algorithms [9].

Neural machine translation is a new approach to machine translation in which a large neural network is trained to maximize translation performance. This is, undoubtedly, a radical departure from existing phrase-based statistical translation approaches, in which a translation system consists of subcomponents that are separately optimized. A bidirectional recurrent neural network (RNN), known as an encoder, is used by the neural network to encode a source sentence for a second RNN, known as a decoder, that is used to predict words in the target language [10].

Now let us compare neural networks with human activity. Whenever a new neural network is created, it is like a child being born. After birth, we start training the network. Unsurprisingly, we may have created the neural networks for certain domains or applications. Here, the difference between neural networks and childbirth is rather obvious. First, a decision to create the needed neural network is made. On the other hand, childbirth results are naturally random. After childbirth, one cannot know whether a child will concentrate on studies through his or her entire life. That is left in the hands of the child and the parents. Parents undoubtedly play an important role in child development, and this aspect is similar to the person creating a neural network. In the same way a child develops to be an expert in a certain domain, neural networks are also trained to be experts in specific domains. Once an automatic learning mechanism is established in neural networks, it practices, and with time it does work on its own, as expected. After proof that the neural network is working correctly, it is called an "expert," as it operates according to its own judgment [11].

Both the human and neural networks can learn and develop to be experts in a certain domain, with both being mortal. But what is the difference between them? The major difference is that humans can forget, unlike neural networks. A neural network will never forget, once trained, as information learned is permanent and hard-coded. Human knowledge may not be permanent. A number of factors may cause the death of brain cells, with stored information getting lost, and we start forgetting [11].

Another major difference is accuracy. Once a particular process is automated via a neural network, the results can be repeated and will definitely remain as accurate as calculated the first time. Humans are rather different. The first ten processes may be accurate, with the next ten characterized by mistakes. Another major difference is speed [11].

**3. Data preparation**

We derived a corpus from the European Medicines Agency (EMEA) parallel corpus with Polish language data included. It was created from EMEA's biomedical PDF documents. The derived corpus includes medical product documents and their translation into 22 official European Union languages [12]. It consists of roughly 1,500 documents for each language, but not all of them are available in every language. The data comprises 80 MB and 1,044,764 sentences constructed from 11.67 million untokenized words [13]. The data is UTF-8 encoded text. In addition, the texts were separated and structured in language pairs. This corpus was chosen as the most similar to medical texts (which could not be accessed in sufficient quantity) in terms of vocabulary and complexity.

The vocabulary consisted of 109,320 unique English and 148,160 unique Polish tokens. When it comes to translation from English to Polish, the disproportionate vocabulary sizes and the number of tokens makes it a challenging task.

Prior to using the training translation model, preprocessing including long sentence (set to 80 tokens) removal to limit model size and computation time. For this purpose, Moses toolkit scripts were employed [14]. Moses is an open source SMT toolkit supporting linguistically-motivated factors, network decoding, and data formats required for efficient use by language and translation models. The toolkit also included an SMT decoder, a wide variety of training tools, tuning tools, and a system applied to a number of translation tasks.

English data preparation was much less complicated as compared to that of Polish data. A tool to clean the English data was developed by eliminating strange symbols, foreign words, etc. The English data had much fewer errors, though there was a need to fix some problems. Translations in languages other than English proved to be problematic, with repetitions, UTF-8 symbols, and unfinished sentences. When corpora are built by automatic tools, such errors are typical.

**4. Evaluation methods**

Human evaluations of machine translation outputs require considerable effort and are quite expensive. They can take a number of days or even weeks for completion. So, it is clear that automatic metrics are required to measure the translation quality from SMT systems. Comparison of SMT translations with human translations is done via different automated metrics. Some of the most widely used SMT metrics include:
- U.S National Institute of Standards & Technology (NIST) metric
- Bilingual Evaluation Understudy (BLEU)
- Translation Error Rate (TER)
- Metric for Evaluation of Translation with Explicit Ordering (METEOR)

Radziszewski has briefly described SMT metrics [15]. BLEU is inexpensive to calculate, is quick to use, correlates well with human evaluation, and is language independent. It is among the most widely-used automated methods of determining machine translation quality. A BLEU score ranges in value from 0 to 1, and it is typically displayed as a percentage. The more the translation correlates with human translation, the closer the score gets to one (100%). In simple words, the BLEU metric is able to measure how many words overlap in a given translation and a reference translation, with sequential words being given higher scores. Scores below 15% indicate that the machine translation engine is unable to provide quality translations, as reported by Lavie [16] and a commercial software manufacturer [17]. A high level of post editing is required for good quality and publishable output translations.

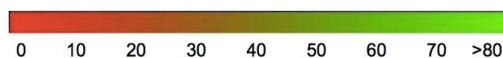

Fig.1. BLEU interpretability scale.

Scores greater than 30% indicate that translations should be understood without any problems. Scores above 50% indicate good quality and fluent translations.

BLEU's general approach [18] is to attempt to match variable phrase lengths to reference translations. The matches/pairs have weighted averages that are, in turn, used for calculation of the metric. A family of BLEU metrics, including standard BLEU, Multi-BLEU, and BLEU-C [19], results from the use of various weighting schemes.

The basic BLEU metric is expressed by the equation [20]:

$$BLEU = P_B \exp\left(\sum_{n=0}^{N} w_n \log p_n\right) \qquad (1)$$

where p_n is an n-gram precision that uses n-grams up to length N and positive weights w_n that sum to one. The brevity penalty P_B is calculated as:

$$P_B = \begin{cases} 1, & c > r \\ e^{(1-\frac{r}{c})}, & c \leq r \end{cases} \quad (2)$$

where c is the length of a candidate translation, and r is the effective reference corpus length [18].

The matches between n-grams of SMT and human translations is calculated by the standard BLEU metric, without consideration of words or phrases within texts. Also, the total count of each SMT candidate word is limited by the human reference translation word count. The BLEU metric avoids bias that would result in the overuse of high-confidence words to boost scores. BLEU applies its text approach to each sentence, and then computes a final score for the overall SMT text output. The geometric mean of individual scores is used to calculate the overall score, with an additional penalty for excessive brevity in translation [18].

The NIST metric, on the other hand, tends to improve the BLEU metric in several ways. It employs the arithmetic mean and the geometric mean of the n-gram matches to emphasize the proper translation of rare words. This metric is improved as compared to the BLEU metric. The NIST metric displays output values ranging from 0 to 15, with a higher value indicating better translation quality [21].

Introduced by the Language Technologies Institute of Carnegie Mellon University, the METEOR metric is also designed to improve the BLEU metric. METEOR was applied without matches for the Polish language. It emphasizes recall by changing the BLEU brevity penalty. Moreover, it considers higher-order n-grams to favor word order matches. It also uses the arithmetic mean vice geometric mean. METEOR is the best option for word-to-word matching. This metric, like BLEU, gives scores ranging from 0 to 100. Doddington provided a detailed description of METEOR metric [22].

Among the most recently-developed SMT metrics is TER. It evaluates the minimum number of human corrections needed for an SMT translation to fully match a reference translation in terms of fluency and meaning. Change of words or phrases, insertion, and removal are some of the human corrections needed. Unlike the other metrics, a lower TER score indicates more similarity with the reference translation. Its scores range from 0 to 100 [23].

## 5. Translation systems

Various experiments have been carried out for evaluation of different versions of translation systems. The experiments included a number of steps including corpora processing, cleaning, tokenization, factorization, splitting, lower casing, and final cleaning. A Moses-based SMT system was used for testing with comparison of performance to a neural network.

The Experiment Management System (EMS) [24] was used together with the SMT Moses Open Source toolkit. In addition, the SRI Language Modelling Toolkit (SRILM) [25] used alongside an interpolated version of Kneser-Ney discounting was employed for training of a 5-gram language model. For word and phrase alignment, the MGIZA++ tool [25], a multithreaded version of the GIZA++ tool, was employed. KenLM [26] was employed to ensure high quality binaries of the language model. Lexical reordering was placed at the mid-bidirectional-fe model, and the phrase probabilities were reordered according to their lexical values. It included three unique reordering orientation types applied to source and target phrases: monotone (M), swap (S), and discontinuous (D). The bidirectional model's reordering includes the probabilities of positions in relation to the actual and subsequent phrases. The probability distribution of English phrases is evaluated by "e," and foreign phrase distribution by "f." For appropriate word alignment, a method of symmetrizing the text was developed. At first, two-way alignments from GIZA++ were structured, which resulted in leaving only the points of alignments appearing in both. The next phase involved combination of additional alignment points appearing in the union. Additional steps contributed to potential point alignment of neighboring and unaligned words. Neighboring can be positioned to the left or right, top or bottom, with an additional diagonal (grow dialog) position. In the final phase, a combination of alignment points between words is considered, where some are unaligned. The application of the grow dialog method will determine points of alignment between two unaligned words [27].

The neural network was implemented using the Groundhog and Theano tools. Most of the neural machine translation models being proposed belong to the encoder-decoder family [28], with use of an encoder and a decoder for every language, or use of a language-specific encoder to each sentence application whose outputs are compared [28]. A translation is the output a decoder gives from the encoded vector. The entire encoder-decoder system, consisting of an encoder and decoder for each language pair, is jointly trained for maximization of the correct translation.

A potential downside to this approach is that a neural network will need to have the capability of compressing all necessary information from a source sentence into a vector of a fixed length. A challenge in dealing with long sentences may arise. Cho et al. showed that an increase in length of an input sentence will result in the deterioration of basic encoder performance [28].

That is the reason behind our encoder-decoder model, which learns to jointly translate and align. After generating a word when translating, the model searches for position sets in the source sentence that contains all the required information. A target word is then predicted by the model based on context vectors [29].

A significant and unique feature of this model approach is that it does not attempt to encode an input sentence into a vector of fixed length. Instead, the sentence is mapped to a vector sequence, and the model adaptively chooses a vector subset as it decodes the translation. This gets rid of the burden of a neural translation model compressing all source sentence information, regardless of length, into a fixed length vector [29].

## 6. Results and conclusions

The experiments were performed to evaluate the optimal translation methods for English to Polish and vice versa. The experiments involved running of a number of tests with use of the developed language data. Random selection was used for data collection, accumulating 1000 sentences for each case. Sentences composed of 50 words or fewer were used, due to hardware limits, with 500,000 training iterations and neural networks having 750 hidden layers. The NIST, BLEU, TER, and METEOR metrics were used for evaluation of the results. The TER metric tool is considered the best one, showing a low value for high quality, while other metrics use high scores to indicate high quality. For comprehension and comparison, all metrics were made to fit in the 0 to 100 range. The results presented in Table 1 are for Polish-to-English and English-to-Polish translation results. Statistical translation results are annotated as SMT in the tables. Translation results from the most popular neural model are annotated as ENDEC, and SEARCH indicates the neural network-trained systems. The results are visualized on the Diagram 1.

Table 1. Polish-to-English and translation results

|        | Polish-to-English |       |        |       | English-to-Polish |       |        |       |
|--------|-------|-------|--------|-------|-------|-------|--------|-------|
| System | BLEU  | NIST  | METEOR | TER   | BLEU  | NIST  | METEOR | TER   |
| SMT    | 36,73 | 55,81 | 60,01  | 60,94 | 25,74 | 43,68 | 58,08  | 53,42 |
| ENDEC  | 21,43 | 35,23 | 47,10  | 47,17 | 15,96 | 31,70 | 62,10  | 42,14 |
| SEARCH | 24,32 | 42,15 | 56,23  | 51,78 | 17,50 | 36,03 | 64,36  | 48,46 |

Diagram 1. Polish-to-English and English-to-Polish translation results

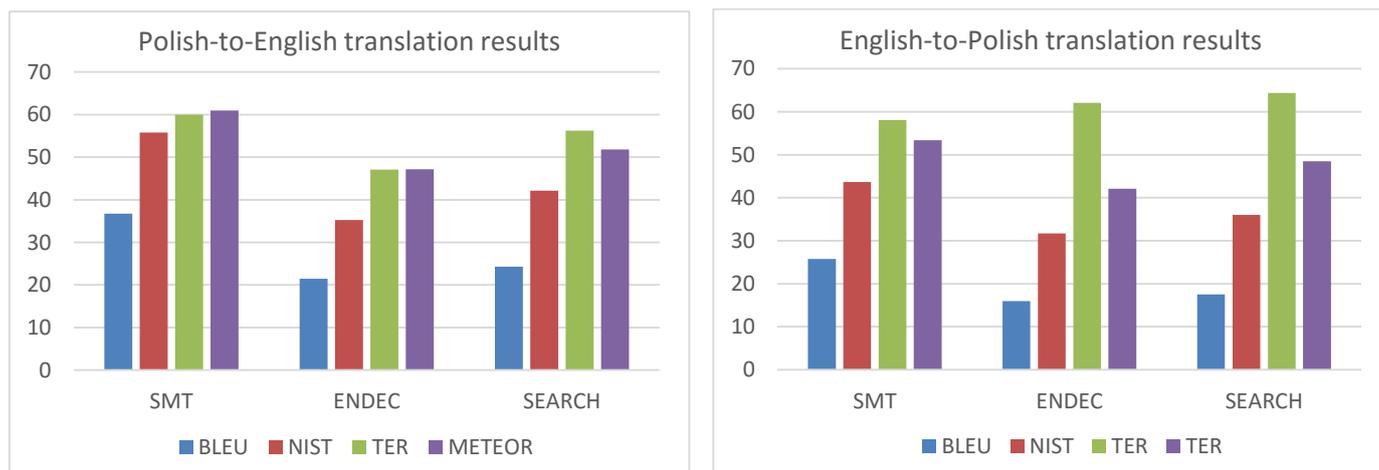

A number of conclusions can be drawn directly from the results of our research experiments. As anticipated, results of the encoder-decoder did not match the SEARCH results. On the other hand, the statistical approach obtained slightly much better results than the SEARCH. Nevertheless, it must be noted that a neural network translation system requires fewer resources for training and maintenance. In addition, the translation results were manually analyzed, due to similarity of the neural network with the brain. In many cases, the neural network substituted words with other words occurring in a similar context. As an example, an input sentence "I can't hear you very well" was first translated to "nie za dobrze pana słyszę" then translated to "słabo pana słyszę" meaning "I hear you poorly." Due to the need to preserve meaning, this leads to a conclusion that an automatic statistical evaluation method is not suited for neural machine translation and also that the final score should be above the one measured. Neural machine translation shows promise for future automatic translation systems. Even though only a few steps have been taken in this research, satisfactory results are already being obtained. With an enlarged number of hidden layers and an increase in the training iteration numbers, language models have the potential to greatly improve in quality [30]. The availability of GPUs can enhance neural network training to become more computationally feasible.

Such systems can also be prepared for any required language pair. Using machine translation to medical texts can have a great potential of ensuring the benefits to patients, including tourists and people who do not know the language of the country in which they require medical help. Improved access to various medical information can be very profitable for patients, medical professionals, and eventually to the medical researchers themselves.

**Acknowledgements**

This work was supported by the CLARIN-PL project.